\documentclass{article}

\usepackage[preprint]{neurips_2025}

\usepackage[utf8]{inputenc} 
\usepackage[T1]{fontenc}    
\usepackage{url}            
\usepackage{amsfonts}       
\usepackage{nicefrac}       
\usepackage{microtype}      
\usepackage[dvipsnames]{xcolor}         
\usepackage{booktabs}       
\usepackage{colortbl}
\usepackage{arydshln}
\usepackage{amsmath}
\usepackage{multirow}
\usepackage{graphicx}   
\usepackage{array} 
\usepackage{amsmath}
\usepackage{adjustbox}
\usepackage{makecell}
\usepackage{wrapfig}

\definecolor{mycolor_blue}{HTML}{E7EFFA}
\definecolor{mycolor_green}{HTML}{E6F8E0}
\definecolor{mycolor_gray}{HTML}{ECECEC}
\definecolor{pearDark}{HTML}{2980B9}
\definecolor{linkcolor}{HTML}{c0392b}
\definecolor{sem}{HTML}{2E75B6}
\definecolor{tok}{HTML}{F3B000}
\usepackage[hidelinks,breaklinks=true,colorlinks,citecolor=Blue,linkcolor=linkcolor]{hyperref} 


\title{TimeHC-RL: Temporal‑aware Hierarchical \\ Cognitive  Reinforcement Learning for Enhancing \\ LLMs' Social Intelligence}

\author{
\textbf{Guiyang Hou}$^{1,*}$ \quad 
\textbf{Xing Gao}$^2$ \quad 
\textbf{Yuchuan Wu}$^2$ \quad 
\textbf{Xiang Huang}$^{2,3}$ \\
\textbf{Wenqi Zhang}$^1$ \quad
\textbf{Zhe Zheng}$^1$ \quad 
\textbf{Yongliang Shen}$^1$ \quad 
\textbf{Jialu Du}$^1$ \\ 
\textbf{Fei Huang}$^2$ \quad 
\textbf{Yongbin Li}$^{2,\dagger}$ \quad 
\textbf{Weiming Lu}$^{1,\dagger}$ \\ 
$^1$ Zhejiang University \quad 
$^2$ Tongyi Lab, Alibaba Group \\
$^3$ Nanjing University \\
\texttt{gyhou@zju.edu.cn, shuide.lyb@alibaba-inc.com, luwm@zju.edu.cn} \\
\\
\url{https://github.com/ZJU-REAL/TimeHC-RL}
}

\begin{document}

\maketitle
\renewcommand{\thefootnote}{\fnsymbol{footnote}}
\footnotetext[1]{~This work was done when the first author was an intern at Tongyi Lab.}
\footnotetext[2]{~Corresponding author.}
\renewcommand{\thefootnote}{\arabic{footnote}}

\begin{abstract}

Recently, Large Language Models (LLMs) have made significant progress in IQ-related domains that require careful thinking, such as mathematics and coding. However, enhancing LLMs' cognitive development in social domains, particularly from a post-training perspective, remains underexplored. Recognizing that the social world follows a distinct timeline and requires a richer blend of cognitive modes (from intuitive reactions (System 1) and
surface-level thinking to deliberate thinking (System 2)) than mathematics, which primarily relies on System 2 cognition (careful, step-by-step reasoning), we introduce \textbf{T}emporal-aware \textbf{H}ierarchical \textbf{C}ognitive \textbf{R}einforcement \textbf{L}earning \textbf{(TimeHC-RL)} for enhancing LLMs' social intelligence. In our experiments, we systematically explore improving LLMs’ social intelligence and validate the effectiveness of the TimeHC-RL method, through five other post-training paradigms and two test-time intervention paradigms on eight datasets with diverse data patterns. Experimental results reveal the superiority of our proposed TimeHC-RL method compared to the widely adopted System 2 RL method. It gives the 7B backbone model wings, enabling it to rival the performance of advanced models like DeepSeek-R1 and OpenAI-O3. Additionally, the systematic exploration from post-training and test-time interventions perspectives to improve LLMs' social intelligence has uncovered several valuable insights.

\end{abstract}    
\section{Introduction}
\label{sec:intro}
Recently, Large Language Models (LLMs) have made significant progress and achieved notable success in IQ-related domains such as mathematics and coding. Several approaches have contributed to this progress: Long-thought Supervised Fine-Tuning (SFT) (even with relatively small data scales, as seen in LIMO~\citep{ye2025limo}), rule-based Reinforcement Learning (RL) (as demonstrated by DeepSeek-R1~\citep{guo2025deepseek} and OpenAI-O1~\citep{jaech2024openai}), and test-time budget forcing (as implemented in s1~\citep{muennighoff2025s1}). However, advancing the cognitive development of LLMs in social domains, despite its significance, has not received sufficient attention and comprehensive exploration.

\begin{figure*}[t!]
    \centering
    \includegraphics[width=0.98\linewidth]{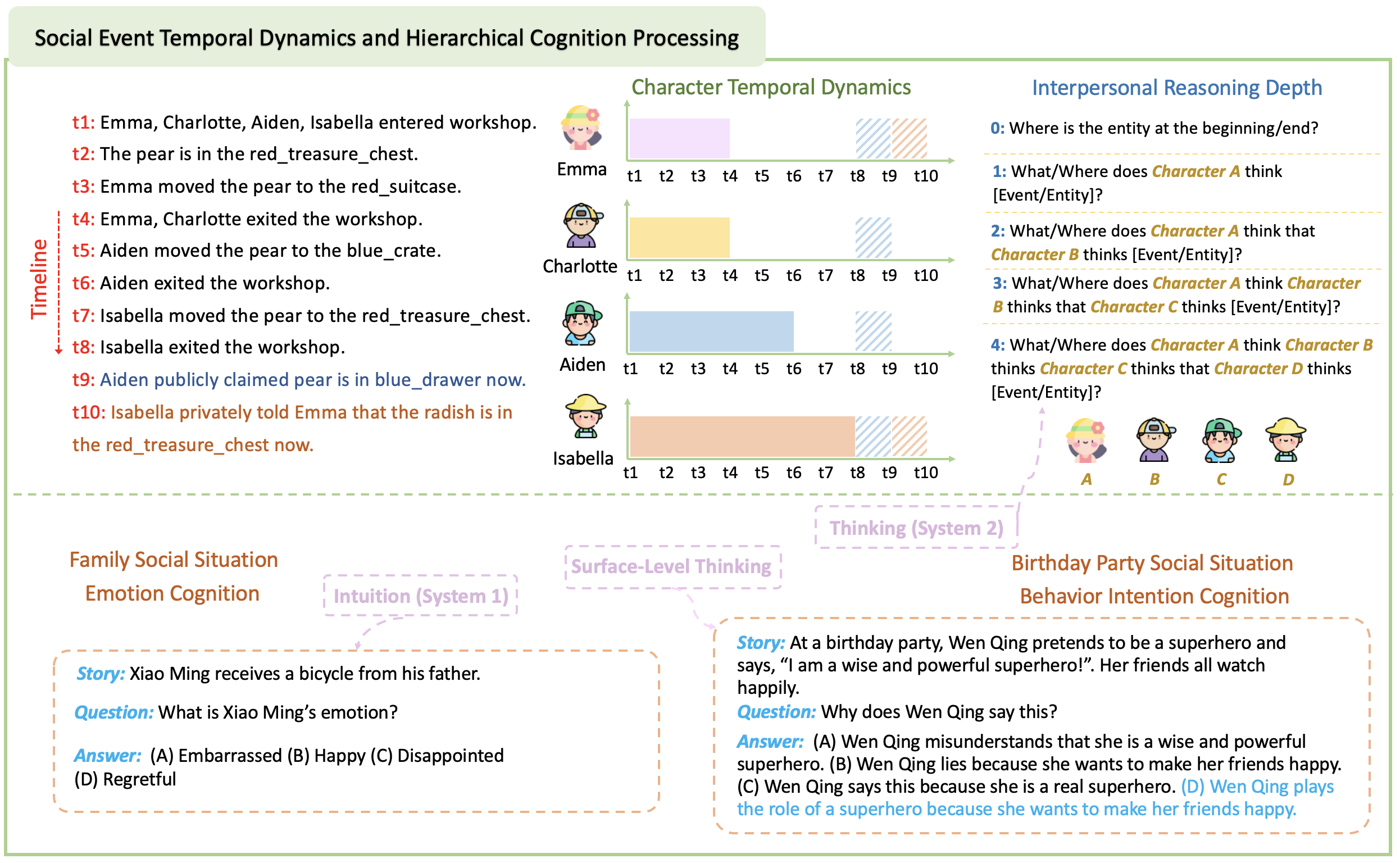}
    \caption{Top: Real-world social events following a clear timeline and character temporal dynamics, interpersonal reasoning presentation. Bottom: Presentation of social situation cognition; Diverse cognitive patterns observed in the social domain.} 
    \label{fig:intro1}
\end{figure*}

Current approaches to enhance LLMs' cognitive performance in the social domain can mainly be divided into three categories: (1) Prompt-based approaches, such as perspective-taking~\citep{wilf2024think, jung2024perceptions}; (2) External tool-based approaches, such as building world models~\citep{huang2024notion}, belief trackers~\citep{sclar2023minding}, and solvers~\citep{hou2024timetom}; (3) Model-based approaches, such as Bayesian models~\citep{zhang2025autotom, shi2025muma, jin2024mmtom}. Despite these advances, \textbf{there remains a notable research gap in systematic exploration from post-training and test-time intervention perspectives.}


We first conduct a comprehensive evaluation and analysis of the advanced DeepSeek-R1 model's performance on social domain benchmarks, specifically ToMBench~\citep{chen2024tombench} and HiToM~\citep{wu2023hi}. Correctly answering questions in ToMBench requires advanced cognition and understanding of social contexts, while correctly answering questions in HiToM requires sophisticated reasoning about interpersonal dynamics in social-event lines. Our experiments show that DeepSeek‑R1 consumes a large number of tokens on both benchmarks; its performance on ToMBench is on par with the GPT-4 series models (78.4\% vs 75.3\%, as seen in Appendix \ref{t1}), while its results on HiToM are noticeably stronger. Based on experimental observations, we analyze the adaptability of DeepSeek-R1's training paradigm in the social domain:
\begin{itemize}
\item Its good performance on HiToM benefits from its superior reasoning capabilities, which are incentivized by rule-based RL. Additionally, the social events in HiToM are relatively basic and simple, requiring lower levels of social situation cognition.

\item Its relatively average performance on ToMBench may be due to a lack of diverse social situations in its training data.

\item It consumes a large number of tokens on both benchmarks. Yet, \textbf{unlike the mathematics domain, where System 2 cognition (careful, step‑by‑step reasoning) is predominant, the social domain involves a richer mix of cognitive modes}: cognition of social situations can be intuitive (System 1) ~\citep{kahneman2011thinking} or involve some basic analytical understanding, while inferring others' mental states may require more deliberate thinking (System 2).

\end{itemize}

In this paper, we introduce \textbf{T}emporal-aware \textbf{H}ierarchical \textbf{C}ognitive \textbf{R}einforcement \textbf{L}earning \textbf{(TimeHC-RL)} for Enhancing LLMs' Social Intelligence. Our key methodological contributions include:

\begin{itemize}
\item \textbf{Addressing real-world temporal dynamics}: Social events and conversations inherently follow temporal sequences with distinct characteristics (Figure \ref{fig:intro1}, Top). Conventional rewards focused merely on format and outcomes prove inadequate for training LLMs to reason effectively across social event timelines and conversation flows. We introduce novel temporal-aware reward mechanisms to address this limitation.

\item \textbf{Implementing hierarchical cognitive processing}: In response to the diverse cognitive patterns observed in the social domain (Figure \ref{fig:intro1}, Bottom), we propose a hierarchical cognition framework that encompasses a spectrum from intuitive reactions (System 1) and surface-level thinking to deliberate thinking (System 2). 
\end{itemize}

In our experiments, we systematically explore improving LLMs' social intelligence and validate the effectiveness of the TimeHC-RL method, through five other post-training paradigms and two test-time intervention paradigms on eight datasets with various data patterns. We use ToMi~\citep{le2019revisiting}, HiToM~\citep{wu2023hi}, ExploreToM~\citep{sclar2024explore}, ToMBench~\citep{chen2024tombench}, SocialIQA~\citep{sap2019social}) as training sets to cultivate LLMs' comprehensive abilities in social situation cognition and sophisticated reasoning about interpersonal dynamics, while using SimpleToM~\citep{gu2024simpletom}, ToMATO~\citep{shinoda2025tomato}, OpenToM~\citep{xu2024opentom}) for Out-Of-Distribution (OOD) evaluation. The experimental results validate the effectiveness of our introduced temporal reward and hierarchical cognitive framework. Our proposed TimeHC-RL method gives the 7B backbone model wings, enabling it to rival the performance of advanced models like DeepSeek-R1 and OpenAI-O3.
Systematic exploration from post-training and test-time interventions perspectives to improve LLMs' social intelligence uncovered several valuable insights:
\begin{itemize}
\item \textbf{SFT memorizes and has limited memory capacity, while RL generalizes.}

\item \textbf{RL implements more effective interpersonal reasoning depth extrapolation.}

\item \textbf{Cognition of social situations cannot be developed through test-time sequential scaling.}

\item \textbf{RL with different cognitive modes shows significant preferences in performance on different types of data.}

\end{itemize}

\section{Dataset Construction}
\label{sec:dataset}


Real‑world social situations are remarkably diverse, and reasoning chains to infer others’ mental states can be deeply layered, for example, questions like 'Where does Bob think Alice thinks...?' (Second-Order Mental Inference, Interpersonal Reasoning Depth 2). To enhance the social capabilities of LLMs, it is crucial to construct dataset that embodies a wide variety of social situations and necessitates sophisticated reasoning of interpersonal dynamics. In this section, we will introduce the dataset we build for LLM Post-Training and Evaluation, while highlighting several exceptional designs. Table \ref{dataset} presents, for each data source, the depth of interpersonal reasoning required by its questions, the extent of real-world cognition demanded, and how the data are utilized for training and evaluation. The specific forms of samples in each data source can be found in Appendix \ref{data source}.




\subsection{Dataset for LLM Post-Training and Evaluation}
\paragraph{Post-Training.} 
We combine data from ToMi~\citep {le2019revisiting}, Hi-ToM~\citep {wu2023hi}, ExploreToM~\citep {sclar2024explore}, ToMBench~\citep {chen2024tombench}, and SocialIQA~\citep {sap2019social} to form our Post-Training dataset. The data format for ToMi, Hi-ToM, and ExploreToM is (Social-Event Lines, Question, Choices), which requires sophisticated reasoning about the interpersonal dynamics in the Social-Event Lines to answer the question correctly. The data format for ToMBench and SocialIQA is mainly (Social Situation, Question, Choices), which requires advanced cognition and understanding of the social situation to answer the question correctly. 

\begin{table}
  \centering
  \caption{ToMi and Hi-ToM's social-event lines are mainly limited to simple object location changes and characters entering and exiting rooms; ExploreToM expands the action space, introducing richer interaction types such as communication between characters and secret observations; while other data sources contain situations, event lines, and conversation flows that are highly aligned with real-world social interactions. Based on the varying degrees of real-world cognitive demands these data sources require, we categorize them into three levels: Basic, Intermediate, and Advanced. }
  %
  \label{dataset}
  \begin{adjustbox}{width=1.0\textwidth}
    \renewcommand{\arraystretch}{1.1}
    \begin{tabular}{cccc}
      \toprule
      \multirow{2}{*}{\textbf{Data Source}}   & \textbf{Real World} & \textbf{Interpersonal} & \multirow{2}{*}{\textbf{Usage}} \\
      & \textbf{Cognition Demand} & \textbf{Reasoning Depth} & \\ 
      \midrule
      ToMi & Basic & 2 & Split for Post-Training (800) and In-Domain Evaluation (200) \\
      \multirow{2}{*}{Hi-ToM} & \multirow{2}{*}{Basic} & \multirow{2}{*}{4} & Reasoning Depth 1,2 for Post-Training (360) \\ & & & Reasoning Depth 3,4 for In-Domain Evaluation (240)\\
      ExploreToM & Intermediate & 2 & Split for Post-Training (2k) and In-Domain Evaluation (300) \\
      ToMBench & Advanced & 1 & Split for Post-Training (2.4k) and In-Domain Evaluation (431) \\
      SocialIQA & Advanced & 1 & Split for Post-Training (2k) and In-Domain Evaluation (120) \\
      \midrule
      SimpleToM & Advanced & 1 & Out-of-Domain Evaluation (120) \\
      OpenToM & Advanced & 2 & Out-of-Domain Evaluation (85) \\
      ToMATO & Advanced & 2 & Out-of-Domain Evaluation (50) \\
      \bottomrule
    \end{tabular}
  \end{adjustbox}
\end{table}

\paragraph{Evaluation.} 
We divide the evaluation into In-Domain evaluation and OOD evaluation. For the In-Domain evaluation, we select non-overlapping data from ToMi, ExploreToM, ToMBench, and SocialIQA that were not used during the Post-Training phase, and Hi-ToM data with interpersonal reasoning depths of 3 and 4. For the OOD evaluation, we select data from SimpleToM~\citep{gu2024simpletom}, OpenToM~\citep{xu2024opentom}, and ToMATO~\citep {shinoda2025tomato}. The ToMATO dataset consists of dialogues between two agents generated with the Sotopia~\citep{zhou2023sotopia} framework; its data format is (Conversation, Question, Choices). OpenToM adopts the same data format as ToMi, Hi‑ToM, and ExploreToM, whereas SimpleToM shares the data format used by ToMbench and SocialIQA. The exceptional designs in our evaluation are summarized below:

\begin{itemize}
\item \textbf{Interpersonal reasoning depth generalization} – In Hi‑ToM, samples with reasoning depths 1 and 2 are used for post‑training, while depths 3 and 4 are reserved for in‑domain evaluation. 

\item \textbf{Inference of agents’ mental states during dialogue interaction} - Using Sotopia-generated agent-agent dialogues, evaluating the ability to infer agents’ mental states during the dialogue interaction. (ToMATO)

\item \textbf{Beyond social‑situation cognition} – Not only assess social-situation cognition but also behavior prediction and judgment. (SimpleToM)
\end{itemize}


%



\section{Methods}
\label{sec:methods}
\subsection{Preliminary}
\paragraph{Group Relative Policy Optimization}
Unlike SFT, which optimizes models through token-level losses, RL-based methods like GRPO utilize policy gradients, calculated from the reward loss, for optimization~\citep{li2025think}. This encourages exploring a much larger solution space~\citep{guo2025deepseek}.

Let $Q$ be the question set, \( \pi_{\theta_{\text{old}}} \) be the policy model and $\{o_1,o_2,\cdots,o_G\}$ be a group of responses from  \( \pi_{\theta_{\text{old}}} \) for a question $q$. Let \( \pi_{\theta_{\text{ref}}} \) denote the frozen reference model. The GRPO algorithms aim to optimize model \( \pi_{\theta} \) by the following objective:
\begin{align*}
    J_{\text{GRPO}}(\theta) &= \mathbb{E}_{q \sim Q, \{o_i\}_{i=1}^G \sim \pi_{\theta_{\text{old}}}} \\
&\Bigg[ 
\frac{1}{G} \sum_{i=1}^G \min \Bigg( 
\frac{\pi_\theta(o_i | q)}{\pi_{\theta_{\text{old}}}(o_i | q)} A_i,
\text{clip}\left(\frac{\pi_\theta(o_i | q)}{\pi_{\theta_{\text{old}}}(o_i | q)}, 1 - \epsilon, 1 + \epsilon\right) A_i 
\Bigg) 
- \beta D_{\text{KL}}(\pi_\theta \| \pi_{\text{ref}}) 
\Bigg],
\end{align*}
where $\epsilon$ and $\beta$ are clipping hyper-parameter
and the coefficient controlling the Kullback–Leibler (KL)
penalty, respectively.
Here, $A_i = \frac{r_i - \text{mean}(\{r_1, r_2, \ldots, r_G\})}{\text{std}(\{r_1, r_2, \ldots, r_G\})}$ 
is the advantage using the group reward $\{r_1, r_2, \ldots, r_G\}$, and $D_{KL}(\pi_\theta \| \pi_{\text{ref}}) = \frac{\pi_{\text{ref}}(o_i | q)}{\pi_\theta(o_i | q)} - \log\left(\frac{\pi_{\text{ref}}(o_i | q)}{\pi_\theta(o_i | q)}\right) - 1
$  is the
KL divergence loss. 
 GRPO eliminates the critic model in PPO by estimating the relative advantage by sampling a group of responses \( \{o_i\}_{i=1}^G \) and normalizing their rewards within the group to compute a relative advantage, which is more computationally efficient~\citep{shao2024deepseekmath}.

\paragraph{Dual-System Theory: Two Modes of Cognitive Processing}
The Dual-System Theory~\citep{sloman1996empirical, kahneman2011thinking, evans2013dual} offers a framework for understanding human cognition. System 1 is characterized by its rapidity, strong intuitive nature, and effortlessness; it handles vast amounts of information in daily life and produces immediate responses. In contrast, System 2 is slow, analytical, and requires conscious attentional investment, a deliberate thinking process that plays a crucial role in complex problem solving.

\subsection{Temporal-aware Hierarchical Cognitive Reinforcement Learning}
\subsubsection{Hierarchical Cognition Framework}
\label{framework}
Cognition of social situations can be intuitive or involve some basic analytical understanding, while inferring others’ mental states may require more deliberate thinking. In light of the diverse cognitive patterns observed in the social domain, we propose a hierarchical cognition framework that ranges from intuitive reaction (System 1), surface-level thinking, to deliberate thinking (System 2). The corresponding behavioral tags for these cognition modes are:

\begin{itemize}

\item \textbf{System1}: <answer> final answer </answer>.

\item \textbf{Surface-level Thinking}: <social context understanding> ... </social context understanding> + <answer> final answer </answer>.

\item \textbf{System2}: <think> thought process </think> + <answer> final answer </answer>.

\end{itemize}

These tags are formulated in alignment with linguistic principles.

\subsubsection{Training Template}
\label{template}
\begin{figure*}[h]
    \centering
    \includegraphics[width=0.9\linewidth]{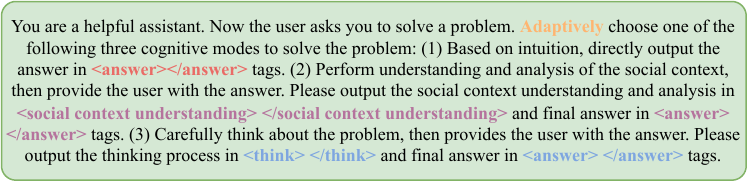}
    \caption{Training template for a model to adaptively choose among three cognitive modes: intuition, surface-level thinking, and deliberate thinking.}
    \label{fig:prompt_case}
\end{figure*}

We train the model to \textbf{adaptively} select one of the three cognitive modes under the hierarchical cognition framework, with the complete content of the training template shown in Figure \ref{fig:prompt_case}.

\subsubsection{Reward Modeling}
The reward function consists of three components: format reward, outcome reward, and temporal reward. The format reward validates that responses adhere to the required structural format, ensuring all elements appear in the correct sequence and are enclosed within appropriate tags:

$$r_{\text{format}} =
\begin{cases} 
1, & \text{if format is correct} \\ 
-1, & \text{if format is incorrect} 
\end{cases}
$$

The outcome reward is a rule-based metric that verifies whether the content enclosed in the <answer> </answer> tags exactly matches the ground truth~(gt) label, which is designed as follows:

$$
r_{\text{accuracy}} =
\begin{cases} 
2, & \text{if answer tag exists and extracted answer matches gt label  } \\ 
-1.5, & \text{otherwise} 
\end{cases}
$$


The temporal reward is a contrastive reward mechanism that explicitly encourages the construction of temporal logic flows. The core idea involves comparing the model's performance on the same social question when social-event lines or conversation flow are provided in two different orders: (1) the temporally ordered sequences, and (2) a shuffled version. For each input question, we generate two groups of responses $\{o_i\}_{i=1}^G$ and $\{\tilde{o_i}\}_{i=1}^{\tilde{G}}$ using the ordered and shuffled inputs, respectively. Let $p$ and $\tilde{p}$ denote the proportion of correct answers in each group. We then define a temporal reward coefficient $r_t$ as:

\begin{equation*}
r_{temporal} = 
\begin{cases}
\alpha, & \text{if } p > \mu \cdot \tilde{p} \\
0, & \text{otherwise}
\end{cases}
\end{equation*}
where $\alpha$ and $\mu$ are hyper-parameters. Here we set $\alpha=0.4$ and $\mu=0.9$. 

This contrastive design incentivizes the model to perform better when the social-event lines or conversation flow is presented in correct temporal order than when it is shuffled. The model receives this positive reinforcement only when its response strategy for a specific question demonstrates clear dependence on temporal information. The temporal reward $r_{temporal}$ is selectively applied only to correct responses, when the model successfully leverages temporal patterns, correct responses receive enhanced reinforcement through this higher reward, while incorrect responses remain unaffected.

The final reward function $r$ is a combination of the three rewards and is defined as:

$$
r = r_{\text{format}} + r_{\text{accuracy}} + r_{\text{temporal}}
$$

\section{Experiments}
\label{sec:experiments}
\subsection{Various Paradigms Setup Used for Systematic Exploration and as Baselines}
\subsubsection{Multiple Post-Training Paradigms}
To verify the effectiveness of the TimeHC-RL method and systematically explore improving LLMs' social intelligence from a post-training perspective, we implement multiple post-training paradigms.
\paragraph{Direct SFT.}
Using social-event lines / social situations and questions as input, with answers as output, to conduct direct SFT on the model.

\paragraph{Long-thought SFT.} 
Using the DeepSeek‑R1‑Distill‑Qwen‑32B model~\citep{guo2025deepseek}, we generate detailed chains of thought for each training sample. After applying basic rule-based filtering to remove low-quality and inconsistent false outputs, we obtain a high-quality long-thought dataset, which is employed for subsequent SFT.    

\paragraph{RL with System 1.}
Unlike RL with System 2 cognition, which encourages models to think deliberately, RL with System 1 cognition prompts models to generate answers intuitively and directly: "Please directly output the answer based on intuition." RL with System 1 cognition eliminates the format reward and retains only the outcome reward.

\paragraph{RL with System 2.} Following~\citet{guo2025deepseek}'s paradigm, we design a prompt that encourages models to engage in a deliberate thinking process before producing the final answer. The prompt is defined as follows: "Please output the deliberate thinking process in <think> </think> and final answer in <answer> </answer> tags, respectively." The reward function consists of two components: format reward and outcome reward.

\paragraph{HC-RL.} Reinforcement learning based on the hierarchical cognitive framework and training template proposed in Section \ref{framework} and \ref{template}.
\subsubsection{Test-Time Intervention: Parallel Scaling and Sequential Scaling}
In addition to the post-training paradigm, we also explore improving LLMs' social intelligence from the perspective of Test-Time Intervention: Parallel Scaling and Sequential Scaling.

\paragraph{Parallel Scaling: Majority Voting.} Repeatedly sample N candidates from the model for each input. From these candidates, select the one that appears most frequently as the final answer~\citep{snell2024scaling}.

\paragraph{Sequential Scaling: Budget Forcing.} To let the model spend more test-time computing on a problem, when the model is about to complete its solution to a problem, append "Wait" to the model's current reasoning trace to encourage the model to engage in more thinking and exploration~\citep{muennighoff2025s1}.

\begin{table}
  \centering
  \caption{Performance evaluation of multiple methods and advanced foundation models in In-Domain scenarios (The HiToM (Third) and HiToM (Fourth) columns also include generalization assessment of interpersonal reasoning depth, because in the post-training phase, we only use interpersonal reasoning problems with reasoning depths of 1 and 2 from HiToM). $\Delta_{\text{Backbone}}$ represents the performance improvement brought by our proposed TimeHC-RL method when applied to the backbone model, and $\Delta_{\text{RL with System 2}}$ represents the performance advantage of our proposed TimeHC-RL method compared to the widely adopted system 2 RL paradigm.}
  
  %
  \label{in domain}
  \begin{adjustbox}{width=1.0\textwidth}
    \renewcommand{\arraystretch}{1.0}
    \begin{tabular}{cccccccc}
      \toprule
      Model   & ToMi & ExploreToM & ToMBench & SocialIQA & HiToM (Third) & HiToM (Fourth) & AVG \\ 
      \midrule
      \multicolumn{8}{c}{\cellcolor{gray!10}BackBone Models} \\
      \midrule
      Qwen2.5-7B-Instruct-1M & 0.60 & 0.45 & 0.69 & 0.77 & 0.29 & 0.26 & 0.51 \\
      \midrule
      \multicolumn{8}{c}{\cellcolor{blue!5}Advanced Foundation Models} \\
      \midrule
      GPT-4o & 0.74 & 0.57 & 0.80 & 0.84 & 0.35 & 0.32 &  0.60 \\
      DeepSeek-R1 & 0.93 & 0.79 & 0.78 & 0.83 & 0.70 & 0.69 & 0.79 \\
      OpenAI-O3 & 0.91 & 0.82 & 0.84 & 0.86 & 0.72 & 0.70 & 0.81 \\
      \midrule
      \multicolumn{8}{c}{\cellcolor{cyan!10}SFT-Based Methods} \\
      \midrule
      Direct SFT & 0.72 & 0.53 & 0.39 & 0.25 & 0.56 & 0.50 & 0.49\\
      Long-thought SFT & 0.89 & 0.74 & 0.73 & 0.63 & 0.55 & 0.42 & 0.66\\
      \midrule
      \multicolumn{8}{c}{\cellcolor{orange!10}RL-Based Methods} \\
      \midrule
        RL with System 1 & 0.84 & 0.93 & 0.79 & 0.81 & 0.60 & 0.53 & 0.75\\
      RL with System 2 & 0.90 & 0.94 & 0.77 & 0.78 & 0.67 & 0.60 & 0.78\\
      HC-RL & 0.92 & 0.94 & 0.81 & 0.79 & 0.65 & 0.64 &  0.79\\
      TimeHC-RL & 0.93 & 0.94 & 0.82 & 0.78 & 0.68 & 0.64 & 0.80\\
      $\Delta_{\text{Backbone}}$ & +0.33 & + 0.49 & + 0.13& + 0.01 & +0.39 &  + 0.38 & +0.29\\
      $\Delta_{\text{RL with System 2}}$ & +0.03 & +0.00 & + 0.05 & + 0.00 & + 0.01 & + 0.04 & +0.02\\
      \midrule
      \multicolumn{8}{c}{\cellcolor{yellow!20}Human Performance} \\
      \midrule
      Human & - & - & 0.85 & 0.84 & - & - & -\\
      \bottomrule
    \end{tabular}
  \end{adjustbox}
\end{table}

\subsection{Implementation Details}
\label{implement}
We use the Qwen2.5-7B-Instruct-1M model~\citep{yang2024qwen2} as the backbone model, which has performance comparable to the Qwen2.5-7B-Instruct model and can handle longer context. We use the TRL~\citep{vonwerra2022trl} and VeRL~\citep{sheng2024hybridflow} frameworks to implement SFT-based methods and RL-based methods, respectively. The specific parameter configurations used for SFT and RL can be found in the Appendix \ref{app: parameter}.  We conduct all experiments on 8 A100 (80GB) GPUs. In addition, we evaluate the performance of the advanced foundation models GPT-4o\footnote{https://openai.com/index/hello-gpt-4o/}, DeepSeek-R1~\citep{guo2025deepseek}, and OpenAI-O3\footnote{https://openai.com/index/introducing-o3-and-o4-mini/} as references. 


\subsection{Metrics and Reliable Reward Signal}
We use the accuracy of question answering as a metric to measure performance. To ensure the reliability and accuracy of the reward signal for stable and effective RL training, we have conducted many detailed data processing steps. For example, in ToMBench, when the answer is given as an option name like "A", we match it with the corresponding specific answer content. Or when the model's response is in the format "A. answer", although it doesn't strictly match the answer, we still consider it correct.

\subsection{Main Results}
As demonstrated in Table \ref{in domain}, our TimeHC-RL method delivers a substantial 29.0 points comprehensive performance improvement over the backbone model in the In-Domain evaluation. It also surpasses the widely adopted System 2 RL paradigm by 2.0 points. Remarkably, with just 7B parameters, our method achieves a comprehensive performance score of 80.0\%, comparable to state-of-the-art models like DeepSeek-R1 and OpenAI-O3. The Direct SFT method and Long-thought SFT method achieve comprehensive performances of 49.0\% and 66.0\%, respectively, showing a notable gap compared to the RL paradigm.

As demonstrated in Table \ref{out of domain}, in the OOD evaluation, we find that during the Post-training phase, our proposed TimeHC-RL method for cultivating social situation cognition and interpersonal reasoning abilities in LLMs brings a 7.0 points improvement, outperforming the System 2 RL paradigm 3.0 points. Meanwhile, SFT-based methods, whether Direct SFT or Long-thought SFT, both reduce the original performance of the backbone model.

{\renewcommand{\arraystretch}{1.04}
\begin{table}
  \centering
  \caption{Performance evaluation of multiple methods and advanced foundation models in OOD scenarios. $\Delta_{\text{Backbone}}$ and $\Delta_{\text{RL with System 2}}$ represents the same meaning as explained in Table \ref{in domain} caption. Due to space constraints, a more comprehensive performance comparison of our method and existing mainstream LLMs can be found in the Appendix \ref{comparison}.}
  %
  \label{out of domain}
  \begin{adjustbox}{width=1.0\textwidth}
    \renewcommand{\arraystretch}{1.0}
    \begin{tabular}{ccccccc}
      \toprule
      Model   & ToMATO (First) & ToMATO (Second)  & SimpleToM (Behavior) & OpenToM (Attitude)  & OpenToM (Location) & AVG \\ 
      \midrule
      \multicolumn{7}{c}{\cellcolor{gray!10}BackBone Models} \\
      \midrule
      Qwen2.5-7B-Instruct-1M & 0.72 & 0.68 & 0.17 & 0.56 & 0.61 & 0.55\\
      \midrule
      \multicolumn{7}{c}{\cellcolor{blue!5}Advanced Foundation Models} \\
      \midrule
      GPT-4o & 0.84 & 0.92 & 0.13 & 0.68 & 0.81 & 0.68 \\
      DeepSeek-R1 & 0.80 & 0.80 & 0.60 & 0.76 & 0.84 & 0.76\\
      OpenAI-O3 & 0.88 & 0.96 & 0.25 & 0.88 & 0.86 & 0.77\\
      \midrule
      \multicolumn{7}{c}{\cellcolor{cyan!10}SFT-Based Methods} \\
      \midrule
      Direct SFT & 0.12 & 0.24 & 0.17 & 0.03 & 0.03 & 0.12  \\
      Long-thought SFT & 0.40 & 0.48 & 0.32 & 0.60 & 0.69 &  0.50 \\
      \midrule
      \multicolumn{7}{c}{\cellcolor{orange!10}RL-Based Methods} \\
      \midrule
      RL with System 1 & 0.64 & 0.64 & 0.22 & 0.60 & 0.68 & 0.56\\
      RL with System 2 & 0.68 & 0.72 & 0.27 & 0.56 & 0.69 & 0.58 \\
     HC-RL & 0.72 & 0.76 & 0.30 & 0.64 & 0.70 & 0.62 \\
      TimeHC-RL & 0.80 & 0.80 & 0.35 & 0.60 & 0.70 & 0.65\\
      $\Delta_{\text{Backbone}}$ & +0.08 & +0.12& +0.18 &+0.04 &+0.09 &  +0.10\\
      $\Delta_{\text{RL with System 2}}$ &+0.12 & +0.08& +0.08 & +0.04 & +0.01 & +0.07\\
      \midrule
      \multicolumn{7}{c}{\cellcolor{yellow!20}Human Performance} \\
      \midrule
      Human & 0.87 & - & - & 0.86 & - & - \\
      \bottomrule
    \end{tabular}
  \end{adjustbox}
\end{table}

Comparing the performance of HC-RL and TimeHC-RL methods in both In-Domain and OOD evaluation scenarios, we find that the introduction of temporal rewards brings performance advantages of 1.0 points and 3.0 points, respectively. This indicates that it can integrate well with the hierarchical cognitive framework, collaboratively enhancing the social intelligence of LLMs.





\subsection{In-Depth Analysis}
\begin{wrapfigure}[15]{r}{0.5\textwidth} 
    \vspace{-5mm} 
    \centering
    \includegraphics[width=0.50\textwidth]{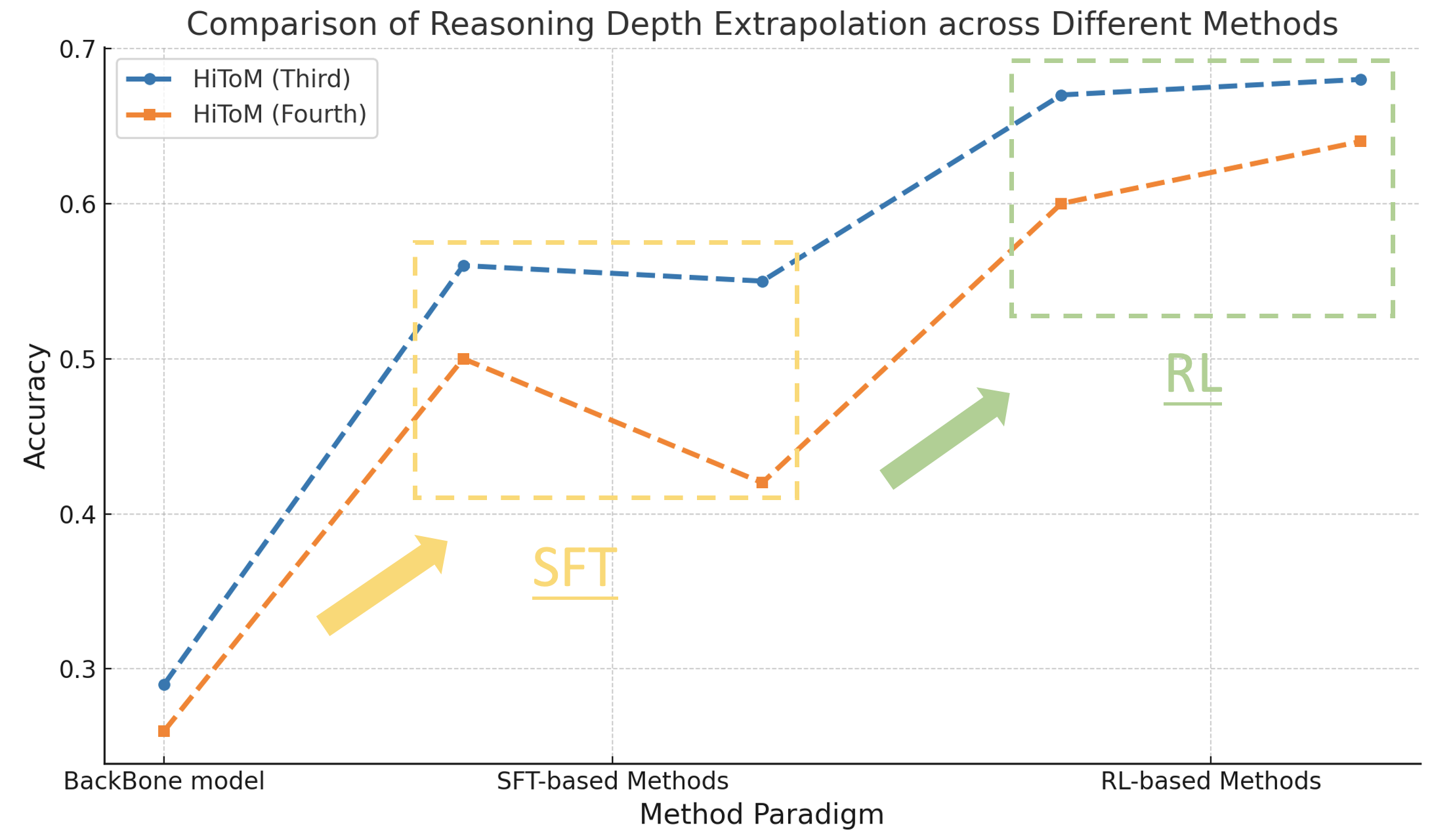}
    \caption{Performance comparison of SFT-based and RL-based methods on interpersonal reasoning depth extrapolation.}
    \label{fig:e1}
\end{wrapfigure}

\paragraph{SFT memorizes, and has limited memory capacity (Direct SFT), while RL generalizes.}
Both direct SFT and long-thought SFT reduce the original performance of the backbone model in OOD evaluation. In contrast, the RL paradigm still provides more or less gains to the backbone model. Furthermore, as shown in Table \ref{in domain}, direct SFT performs relatively well on ToMi, ExploreToM, HiToM (Third), and HiToM (Fourth), which are datasets focusing on interpersonal reasoning, but performs poorly on ToMBench and SocialIQA, which focus on social situation cognition. This indicates that the direct SFT method has lower tolerance for data patterns, which is very unfavorable for the development of social intelligence, considering the inherent complexity of social intelligence. However, the long-thought SFT method improves this aspect, demonstrating better memory capacity.


\paragraph{RL implements more effective interpersonal reasoning depth extrapolation.}
Compared to direct SFT and long-thought SFT methods, although all methods underwent post-training on interpersonal reasoning problems with reasoning depths of 1 and 2, the RL method significantly outperforms on interpersonal reasoning problems with reasoning depths of 3 and 4 (0.68, 0.64), substantially surpassing both direct SFT (0.56, 0.50) and long-thought SFT (0.55, 0.42) methods, as shown in Figure \ref{fig:e1}.


\paragraph{Cognition of social situations cannot be developed through test-time sequential scaling.}

As shown in Table \ref{scaling}, we explore the utility of Majority Voting (Parallel Scaling) and Budget Forcing (Sequential Scaling) in the social domain. We find that the application of the majority voting strategy did not demonstrate any clearly capturable characteristics, whether focusing on advanced cognition of social situations or interpersonal reasoning data. Budget forcing shows gains for data focusing on interpersonal reasoning, but had no effect on data focusing on advanced cognition of social situations. We speculate that developing social situational cognition may necessarily require either incorporating diverse social situations in the training data or increasing the model size.

\paragraph{RL with different cognitive modes shows significant preferences in performance on different types of data.}
As shown in Table \ref{in domain}, careful observation of the performance of RL with System 2 versus RL with System 1 on ToMi, ExploreToM, ToMBench, and SocialIQA reveals that RL with System 2 performs better on ToMi and ExploreToM datasets that focus on interpersonal reasoning, while RL with System 1 performs better on ToMBench and SocialIQA datasets that focus on social situational cognition. This further demonstrates the necessity of building a hierarchical cognitive framework to develop social intelligence in LLM. In Appendix \ref{app: case}, we present how LLM adaptively employs different cognitive modes to address various data types in the social domain.

{\renewcommand{\arraystretch}{1.04}
\begin{table}
  \centering
  \caption{Performance evaluation of applying majority voting strategy to the Qwen2.5-7B-Instruct-1M backbone model, as well as applying budget forcing to the Long-thought SFT model.}
  %
  \label{scaling}
  \begin{adjustbox}{width=1.0\textwidth}
    \renewcommand{\arraystretch}{1.0}
    \begin{tabular}{ccccccc}
      \toprule
      Model   & ToMi & HiToM (Third) & HiToM(Fourth) & ToMBench (Moral Emotion)  & ToMATO & AVG \\ 
      \midrule
      \multicolumn{7}{c}{\cellcolor{blue!5}Majority Voting Parallel Scaling} \\
      \midrule
      Qwen2.5-7B-Instruct-1M & 0.60 & 0.29 & 0.26 & 0.65 & 0.68 & 0.50\\
      Majority Voting (N = 8) & 0.71 & 0.22 & 0.20 & 0.75 & 0.56 & 0.49\\
      $\Delta$ & +0.11 & -0.07 & -0.06 & +0.10 & -0.12 & -0.01\\
      \midrule
      \multicolumn{7}{c}{\cellcolor{orange!10}Budget Forcing Sequential Scaling} \\
      \midrule
      Long-thought SFT & 0.89 & 0.55 & 0.42 & 0.70 & 0.40 & 0.60\\
      Budget Forcing (M = 1) & 0.90 & 0.58 & 0.49 & 0.68 & 0.44 & 0.62\\
      $\Delta$ & +0.01 & +0.03 & +0.07 & -0.02 & +0.04 & +0.02\\
      \bottomrule
    \end{tabular}
  \end{adjustbox}
\end{table}

\section{Conclusions, Limitations and Future Works}
\label{limit}
In this paper, considering the temporal dynamics of real-world social events and that the social domain involves a richer mix of cognitive modes (from intuitive reaction, surface-level thinking, to deliberate thinking), we introduce \textbf{T}emporal-aware \textbf{H}ierarchical \textbf{C}ognitive \textbf{R}einforcement \textbf{L}earning \textbf{(TimeHC-RL)} to enhance LLMs' social intelligence. We obtain a 7B parameter model that demonstrates strong comprehensive capabilities in social situation cognition and interpersonal reasoning, performing well across multiple social domain benchmarks. In our experiments, we systematically explore improving LLMs' social intelligence and validate the effectiveness of the TimeHC-RL method, through five other post-training paradigms and two test-time intervention paradigms on eight datasets with diverse data patterns, revealing several valuable insights that lay the foundation for future research on social intelligence in LLMs. Some limitations and potential future works are listed as follows:
\begin{itemize}
\item \textbf{Beyond Situational Intelligence and Cognitive Intelligence} In our paper, we focus more on situational intelligence and cognitive intelligence. The ability to behave and interact (behavioral intelligence) is also very important.

\item \textbf{Scalable Social Situation Framework} We believe that incorporating richer social situations in training data, exposing LLMs to a more diverse social world, is very helpful for enhancing the social intelligence of LLMs. Therefore, forming a scalable social situation framework is extremely important.

\item \textbf{Experiments with multiple model sizes} In our paper, we only conduct experiments with a 7B parameter model. Considering that models of different sizes have different inherent knowledge levels, and larger models have higher cognitive capacity, experiments with multiple model sizes might reveal more valuable insights for improving the social intelligence of LLMs.
\end{itemize}

\bibliographystyle{plainnat}
\bibliography{refs}



\appendix
\section{Technical Appendices and Supplementary Material}
\subsection{Preliminary Experiments——DeepSeek-R1's Evaluation Performance on ToMBench}
\label{t1}
We evaluate the performance of the DeepSeek-R1 model on ToMBench, and compare it with models from the GPT-4~\citep{gpt4} series, Claude series~\citep{claude3.5sonnet}, and Qwen-Max~\citep{qwenmax} models. The comparison results are shown in Table \ref{test}.

\subsection{Related Work}
\paragraph{Strategies for Enhancing LLMs' Cognitive Development in Social Domain}
\label{t2}
To enhance LLMs' cognitive development in the social domain, existing methods can be mainly divided into three categories: (1) Prompt-based Methods (2) Tool-based Methods (3) Model-based Methods. SimToM~\citep{wilf2024think} prompts LLMs to adopt perspective-taking cognitive strategies, while PercepToM~\citep{jung2024perceptions} improves perception-to-belief inference by extracting relevant contextual details. Meanwhile, ~\citet{huang2024notion} utilizes an LLM as a world model to track changes in environmental entity states and character belief states.  ~\citet{hou2024timetom} proposes a belief solver that transforms higher-order social cognition problems into lower-order social cognition problems based on intersections over time sets, while SymbolicToM~\citep{sclar2023minding} uses graphical representations to track characters' beliefs. Additionally, AutoToM, MMToM, and MuMA-ToM~\citep{zhang2025autotom, shi2025muma, jin2024mmtom} propose Bayesian model-based methods. However, \textbf{there remains a notable research gap in systematic exploration from post-training and test-time intervention perspectives.}


\begin{table}
  \centering
  \caption{DeepSeek-R1's evaluation performance on ToMBench, where UOT represents Unexpected Outcome Test, SIT represents Scalar Implicature Task, PST represents Persuasion Story Task, FBT represents False Belief Task, AST represents Ambiguous Story Task, HT represents Hinting Test, SST represents Strange Story Task, FRT represents Faux-pas Recognition Test.}
  \label{test}
  \begin{adjustbox}{width=1.0\textwidth}
    \renewcommand{\arraystretch}{1.0}
    \begin{tabular}{cccccccccc}
      \toprule
      Model   & UOT & SIT & PST & FBT & AST & HT & SST & FRT & AVG \\ 
      \midrule
        GPT-4-0613 & 0.713 & 0.49 & 0.58 & 0.863 & 0.84 & 0.796 & 0.83 & 0.766 & 0.735 \\
        GPT-4-1106 & 0.767 & 0.48 & 0.61 & 0.908 & 0.83 & 0.883 & 0.762 & 0.786 & 0.753 \\
        Claude-3.5-Sonnet-20240620 & 0.733 & 0.505 & 0.61 & 0.858 & 0.895 & 0.971 & 0.771 & 0.834 & 0.772 \\
        Claude-3.5-Sonnet-20241022 & 0.78 & 0.615 & 0.61 & 0.88 & 0.875 & 0.961 & 0.862 & 0.83 & 0.802 \\
        Qwen-Max-0919 & 0.74 & 0.475 & 0.62 & 0.898 & 0.815 & 0.874 & 0.813 & 0.802 & 0.755 \\
        DeepSeek-R1 & 0.797 & 0.565 & 0.56 & 0.895 & 0.845 & 0.951 & 0.848 & 0.809 & 0.784 \\
        Human & \textbf{0.893} & \textbf{0.755} & \textbf{0.70} & \textbf{0.868} & \textbf{0.95} & \textbf{0.971} & \textbf{0.892} & \textbf{0.804} & 0.854 \\
      \bottomrule
    \end{tabular}
  \end{adjustbox}
\end{table}

\paragraph{LLM Post-Training}
Supervised Fine-Tuning (SFT) and Reinforcement Learning (RL) have been widely used in LLM Post-Training to improve performance on specific tasks.
There are also multiple studies exploring and analyzing these two different Post-Training methods, such as "What LLMs Can—and Still Can't—Solve after SFT?"~\citep{hou2024progressive, chen2025sft, sun2025climbing} and "SFT Memorizes, RL Generalizes" observations~\citep{chu2025sft}. Rule-based RL has already been widely applied to multiple domains beyond mathematics~\citep{hu2025open} and coding~\citep{wei2025swe}, such as image classification, emotion classification tasks~\citep{li2025think, he2025gencls++}, search engine calling~\citep{jin2025search, song2025r1}, video reasoning~\citep{feng2025video}, logic puzzles~\citep{xie2025logic}, machine translation~\citep{feng2025mt}, and more~\citep{li2025relation, xia2025gui, zhou2025sweet}.





\paragraph{Test-Time Scaling}
Test-time scaling methods can be divided into 1) Sequential, where 
later computations depend on earlier ones (e.g., a long reasoning trace), and 2) Parallel, which relies on multiple solution attempts generated in parallel and selecting the best via majority voting or reward model (process-based or outcome-based)~\citep{snell2024scaling, brown2024large, liu2024don, huang2024queryagent, wang2024helpsteer2, zeng2024scaling, qi2024mutual}. s1 in the mathematics domain, m1~\citep{huang2025m1} in the medical domain, and Z1~\citep{yu2025z1} in the code domain are all recent research works related to test-time scaling~\citep{muennighoff2025s1}. The "budget forcing" proposed in s1 refers to appending a "Wait" token to the model's current reasoning trace to encourage the model to engage in more thinking and exploration. In m1, it is precisely about applying budget forcing to the medical domain.





\subsection{Training Parameter Configurations}
\label{app: parameter}
\paragraph{SFT-based Methods.}  The SFT training process employs full-parameter fine-tuning with DeepSpeed ZeRO-3 optimization~\citep{rajbhandari2020zero}. We conduct three epochs of training in bfloat16 precision, with a learning rate of 5e-5, a per-device train batch size of 1, and a cutoff length of 16384. We save the model once every 500 training steps. 

\paragraph{RL-based Methods.} 
The RL training process uses a train batch size of 8, with the maximum input prompt length set to 1536 and the maximum response length set to 2048. The learning rate is set to 3e-7, and the KL loss coefficient is set to 0.001 to ensure sufficient optimization of the policy model. The temperature coefficient is set to 1.0. In the GRPO algorithm, the number of samples in each group is set to 8. The model is saved every 500 steps, and a validation evaluation is performed every 25 steps.

\subsection{Case Study}
\label{app: case}
In the Figure \ref{fig:case study}, we present how LLM adaptively employs different cognitive modes to address various data types in the social domain. For question with interpersonal reasoning depth of 4, LLM adopts the cognitive mode of <think> + <answer> to solve. For a simple social situational cognition question, LLM adopts the cognitive mode of <answer> or <social context understanding> + <answer> to solve.
\begin{figure}
    \centering
    \includegraphics[width=\linewidth]{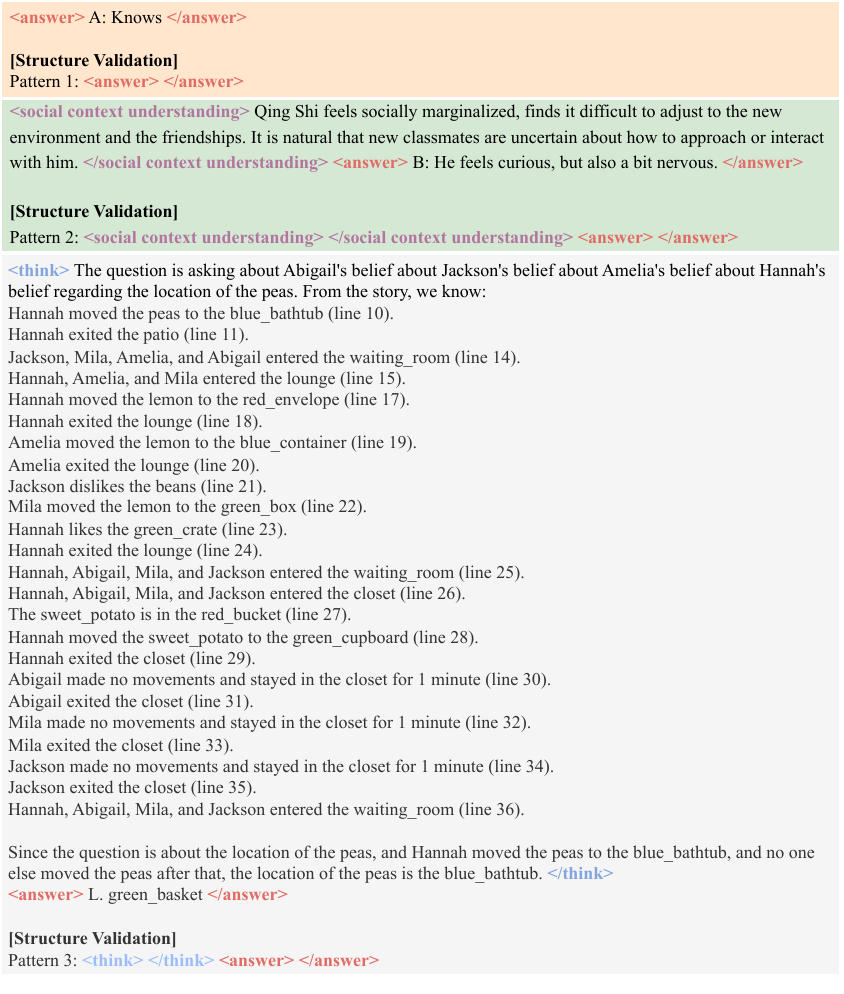}
    \caption{LLM adaptively employs different cognitive modes to address social situation cognition and interpersonal reasoning question in the social domain.}
    \label{fig:case study}
\end{figure}

\subsection{Sample Examples from Data Source}
\label{data source}
For the data sources we use, we present sample examples from Figure \ref{fig:hitom} to Figure \ref{fig:tomatos}. Figure \ref{fig:hitom} presents a sample example from the HiToM data source, where the Story consists of Event lines, with its notable characteristic being the inclusion of interpersonal reasoning questions with deeper reasoning depth. Figure \ref{fig:exploretom} presents a sample example from the ExploreToM data source, whose Story includes some advanced socio-cognitive events, such as "told privately," "witnessed this action in secret," "got distracted," etc. Figures \ref{fig:tombench} and \ref{fig:socialqa} present sample examples from the ToMBench and SocialIQA data sources, respectively, examining models' cognition of social situations, with the data format being (social situation, question, choices). Figure \ref{fig:simpletom} presents a sample example from the SimpleToM data source, which not only assesses social-situation cognition but also behavior judgment. Figure \ref{fig:opentom} presents sample examples from the OpenToM data source, which includes Location questions and Attitude questions. Figures \ref{fig:tomatof} and \ref{fig:tomatos} present sample examples from the ToMATO data source with reasoning depths of 1 and 2, respectively. Using Sotopia-generated agent-agent dialogues, ToMATO evaluates the LLMs' ability to infer agents' mental states during dialogue interaction.

\begin{figure}
    \centering
    \includegraphics[width=1.0\linewidth]{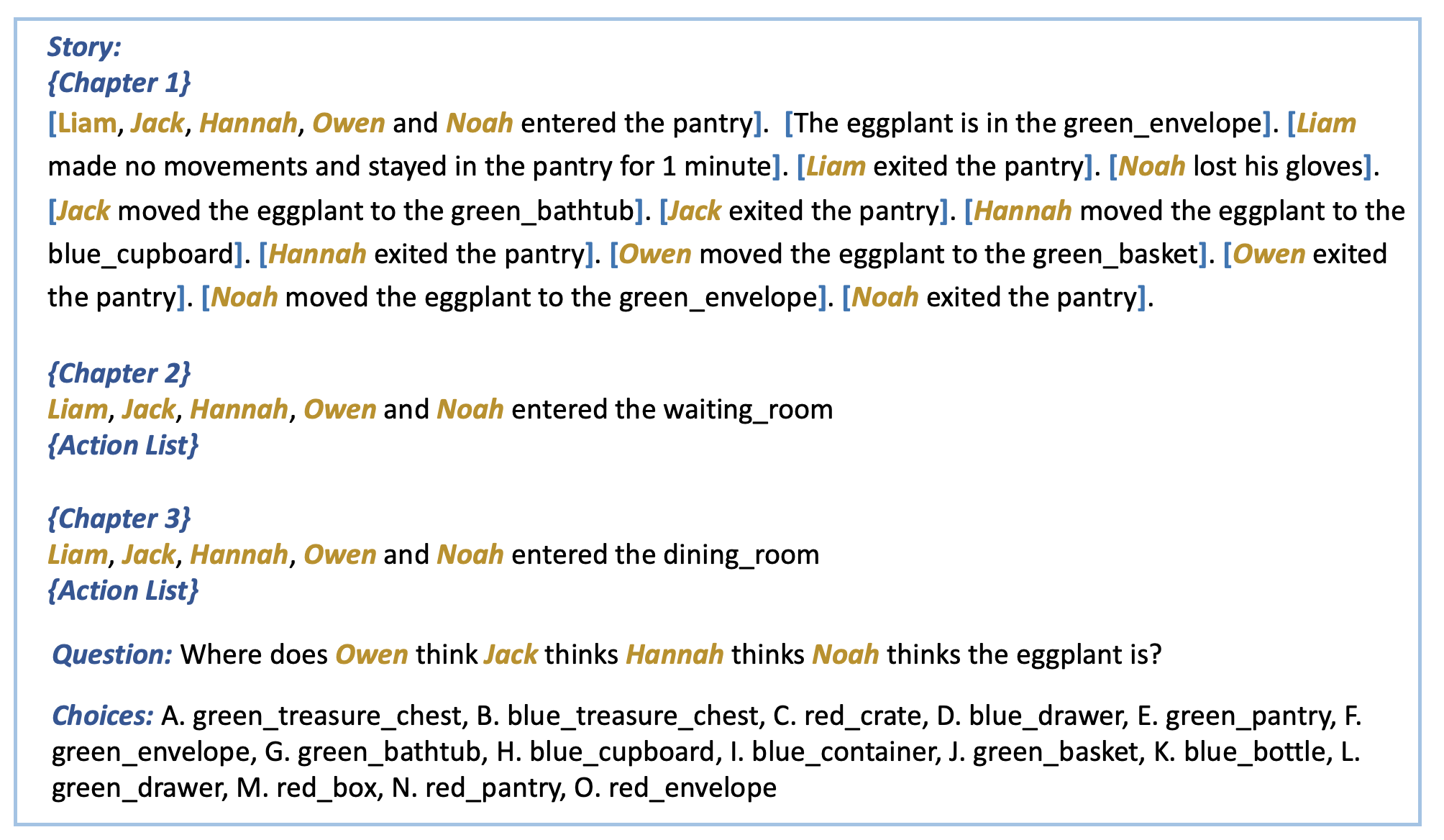}
    \caption{Sample example from HiToM. Data format: (Social-Event Lines, Question, Choices).}
    \label{fig:hitom}
\end{figure}

\begin{figure}
    \centering
    \includegraphics[width=1.0\linewidth]{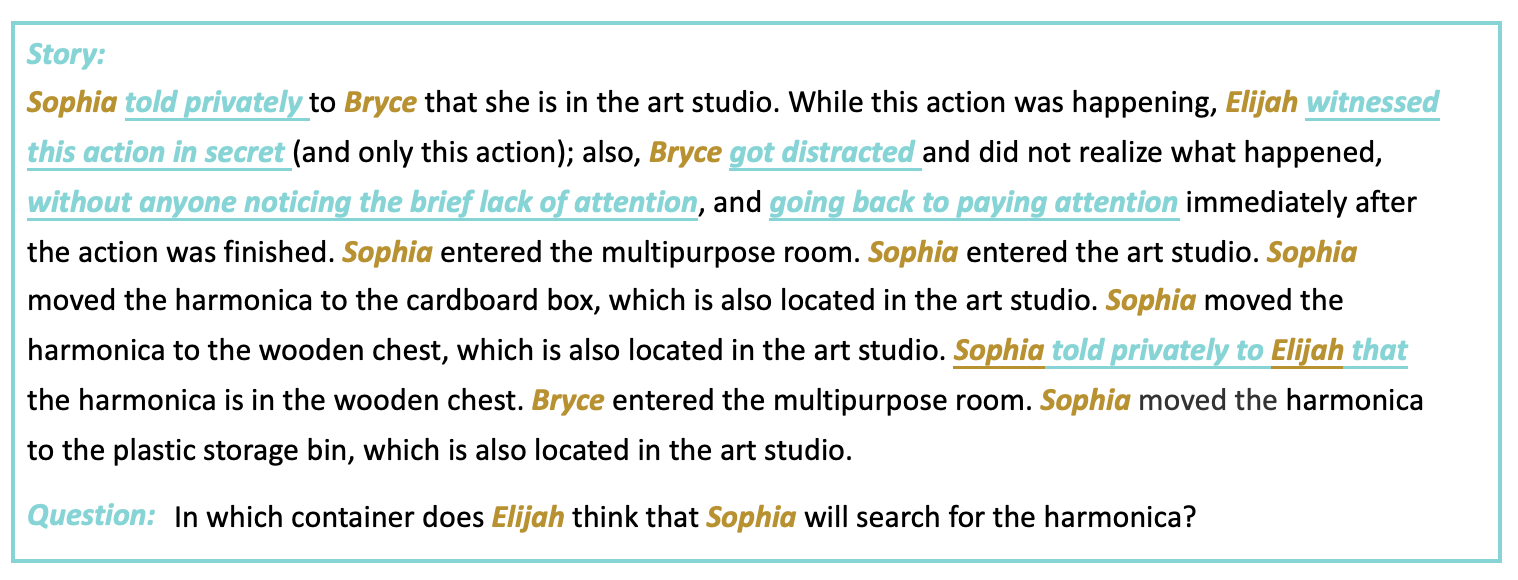}
    \caption{Sample example from ExploreToM. Data format: (Social-Event Lines, Question).}
    \label{fig:exploretom}
\end{figure}

\clearpage
\begin{figure}
    \centering
    \includegraphics[width=\linewidth]{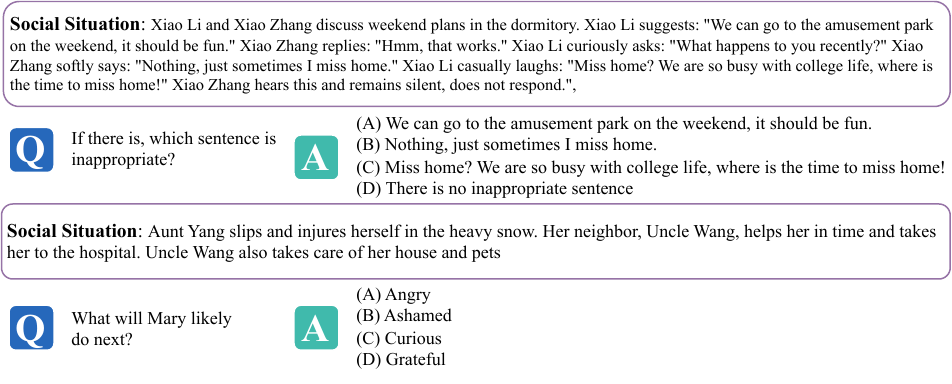}
    \caption{Sample example from ToMbench. Data format: (Social Situation, Question, Choices).}
    \label{fig:tombench}
\end{figure}

\begin{figure}
    \centering
    \includegraphics[width=\linewidth]{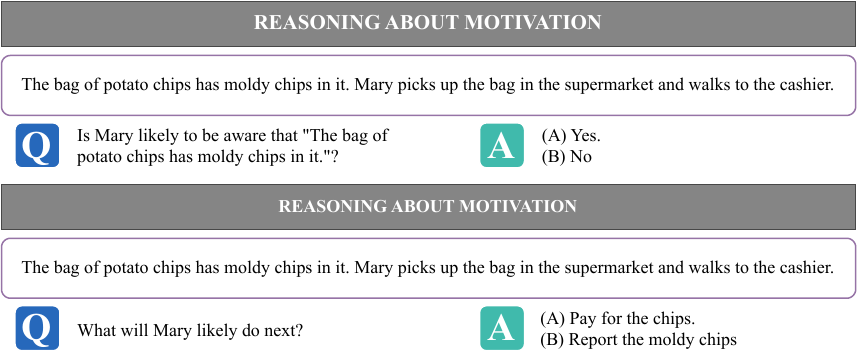}
    \caption{Sample example from SimpleToM, SocialIQA. Data format: (Social Situation, Question, Choices). }
    \label{fig:socialqa}
\end{figure}

\begin{figure}
    \centering
    \includegraphics[width=\linewidth]{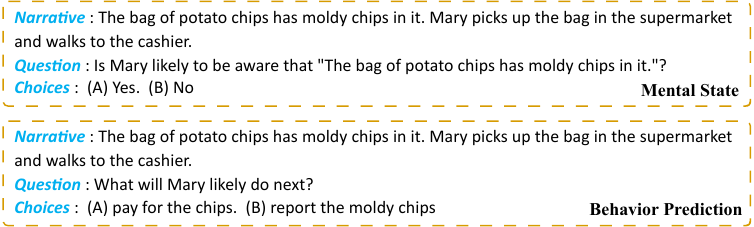}
    \caption{Sample example from SimpleToM, SocialIQA. Data format: (Social Situation, Question, Choices). }
    \label{fig:simpletom}
\end{figure}

\begin{figure}
    \centering
    \includegraphics[width=\linewidth]{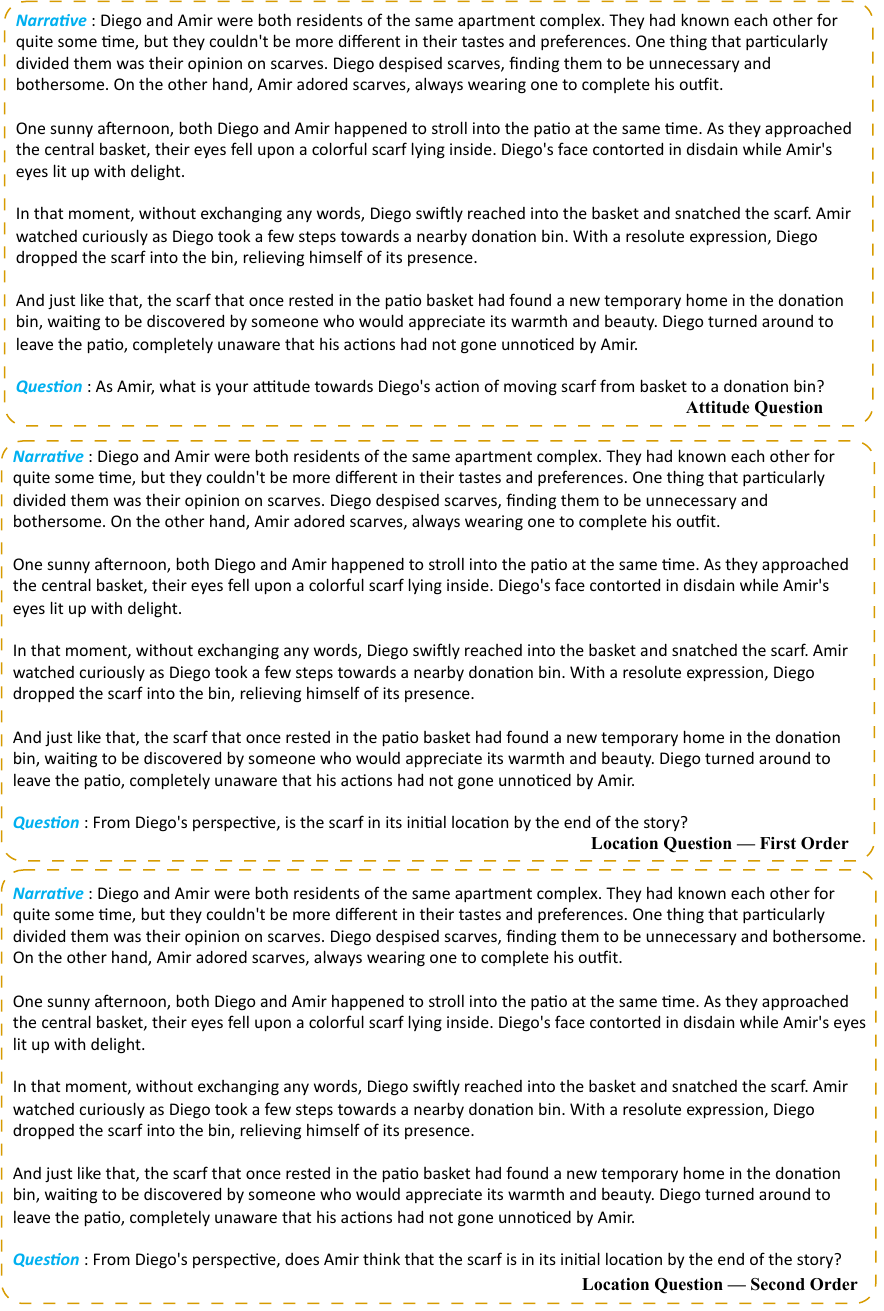}
    \caption{Sample example from OpenToM. Data format: (Social Situation, Question, Choices).}
    \label{fig:opentom}
\end{figure}

\begin{figure}
    \centering
    \includegraphics[width=\linewidth]{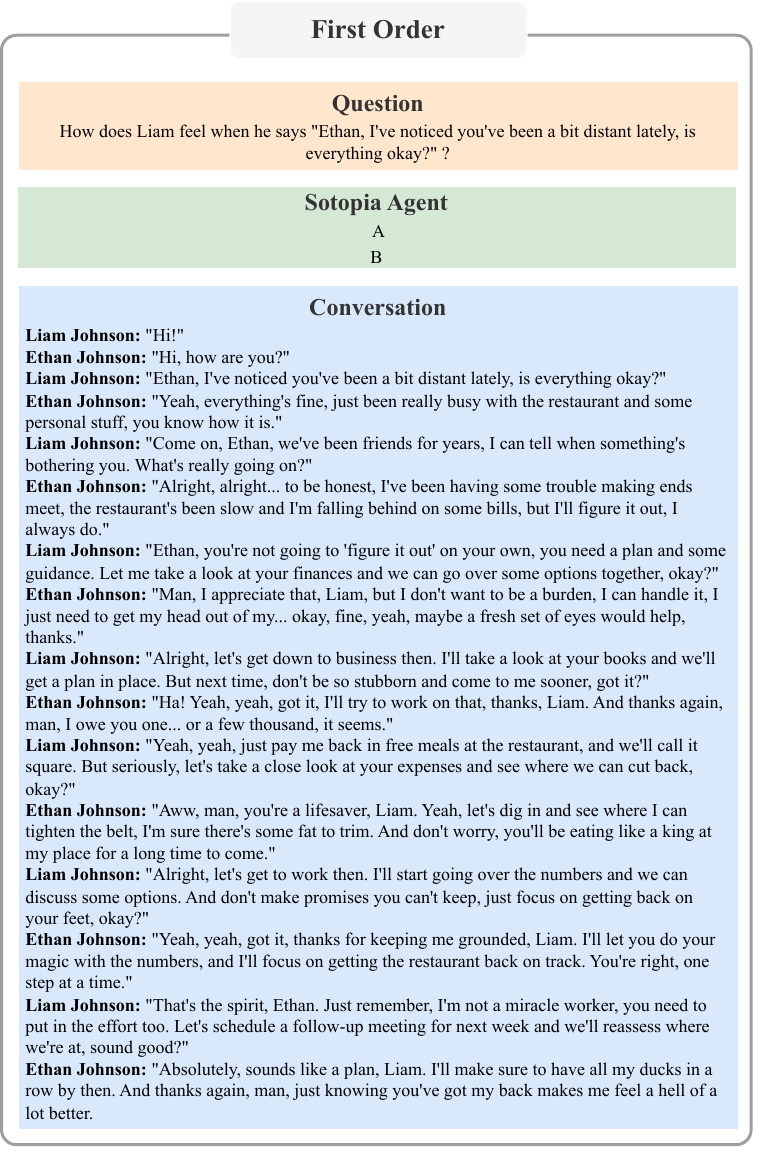}
    \caption{Sample example from ToMATO (First order). Data format: (Conversation, Question, Choices).}
    \label{fig:tomatof}
\end{figure}

\begin{figure}
    \centering
    \includegraphics[width=\linewidth]{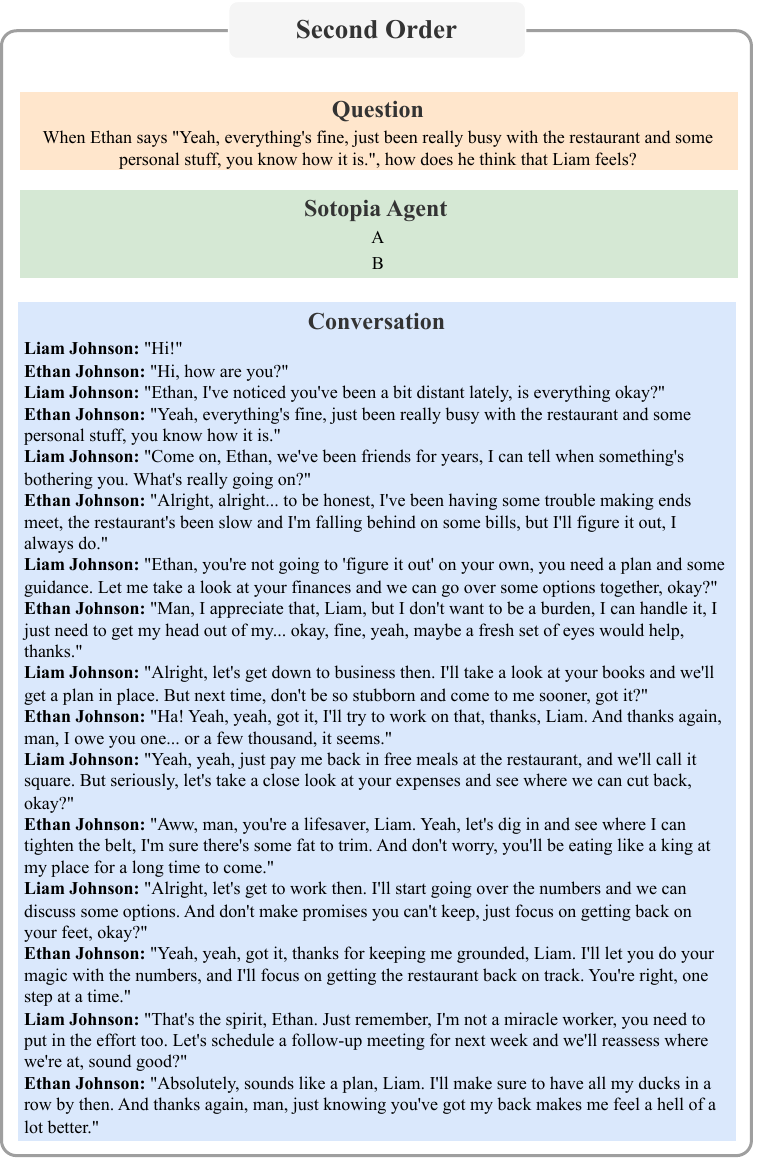}
    \caption{Sample example from ToMATO (Second order). Data format: (Conversation, Question, Choices).}
    \label{fig:tomatos}
\end{figure}

\subsection{Comprehensive Comparison of Our Method and Existing LLMs}
\label{comparison}
Through extensive literature research, we have collected as much data as possible on the performance of existing LLMs on corresponding In-Domain and Out-of-Domain data sources, and conduct a comprehensive comparison with the performance of our proposed TimeHC-RL method (applied to 7B models) as shown in Table \ref{baseline_1} and \ref{baseline_2}.

\begin{table}
  \centering
  \caption{A comprehensive performance comparison of our method and existing LLMs on several In-Domain data sources}
  \label{baseline_1}
  \begin{adjustbox}{width=1.0\textwidth}
    \renewcommand{\arraystretch}{1.0}
    \begin{tabular}{ccccccc}
      \toprule
      Model   & ToMi & ExploreToM & ToMBench & SocialIQA & HiToM (Third) & HiToM (Fourth) \\ 
      \midrule
        ChatGLM3-6B & - & - & 0.43 & - & - & - \\
        Llama2-13B-Chat & - & - & 0.43 & - & - & - \\
        Llama3.1-8B-Instruct & 0.68 & - & - & - & - & - \\
        Llama3.1-70B-Instruct & - & 0.48 & - & - & - & - \\
        Baichuan2-13B-Chat & - & - & 0.49 & - & - & - \\
        Guanaco-65B & - & - & - & - & 0.08 & 0.06 \\
        Claude-instant & - & - & - & - & 0.08 & 0.07 \\
        Mistral-7B & - & - & 0.49 & - & - & - \\
        Mixtral-8x7B & - & - & 0.54 & - & - & - \\
        Mixtral-8x7B-Instruct & - & 0.47 & - & - & - & - \\
        Qwen2.5-7B-Instruct & 0.67 & - & - & - & - & - \\
        Qwen-14B-Chat & - & - & 0.58 & - & - & - \\
        GPT-3.5-Turbo & - & - & - & - & 0.03 & 0.01 \\
        GPT-3.5-Turbo-0613 & - & - & 0.58 & - & - & - \\
        GPT-3.5-Turbo-1106 & - & - & 0.59 & - & - & - \\
        GPT-4-32K & - & - & - & - & 0.18 & 0.15 \\
        GPT-4-0613 & - & - & 0.71 & - & - & - \\
        GPT-4-1106 & - & - & 0.74 & - & - & - \\
        GPT-4o & - & 0.47 & - & - & - & - \\
        TimeHC-RL (7B) & \textbf{0.93} & \textbf{0.94} & \textbf{0.82} & \textbf{0.78} & \textbf{0.68} & \textbf{0.64} \\
      \bottomrule
    \end{tabular}
  \end{adjustbox}
\end{table}

\begin{table}
  \centering
  \caption{A comprehensive performance comparison of our method and existing LLMs on several Out-of-Domain data sources}
  \label{baseline_2}
  \begin{adjustbox}{width=1.0\textwidth}
    \renewcommand{\arraystretch}{1.0}
    \begin{tabular}{cccccc}
      \toprule
      Model   & ToMATO(First) & ToMATO(Second)  & SimpleToM(Behavior) & OpenToM(Attitude)  & OpenToM(Location) \\ 
      \midrule
        Llama2-Chat-7B & - & - & - & 0.24 & 0.37 \\
        Llama2-Chat-13B & - & - & - & 0.37 & 0.37 \\
        Llama2-Chat-70B & - & - & - & 0.41 & 0.34 \\
        Llama3-8B & 0.54 & 0.40 & - & - & - \\
        Llama3-70B & 0.81 & 0.71 & - & - & - \\
        Llama3.1-8B & 0.64 & 0.46 & \textbf{0.54} & - & - \\
        Llama3.1-70B & \textbf{0.82} & 0.73 & - & - & - \\
        Llama3.1-405B & - & - & 0.10 & - & - \\
        Gemma2 & 0.79 & 0.71 & - & - & - \\
        Claude-3-Kaiku & - & - & 0.17 & - & - \\
        Claude-3-Opus & - & - & 0.10 & - & - \\
        Claude-3.5-Sonnet & - & - & 0.25 & - & - \\
        Mistral-7B & 0.65 & 0.56 & - & - & - \\
        Mixtral-8x7B & 0.65 & 0.57 & - & - & - \\
        Mixtral-8x7B-Instruct & - & - & - & 0.40 & 0.47 \\
        GPT-3.5 & - & - & 0.29 & - & - \\
        GPT-3.5-Turbo & 0.60 & 0.51 & - & 0.38 & 0.41 \\
        GPT-4 & - & - & 0.20 & - & - \\
        GPT-4-Turbo & - & - & - & 0.54 & 0.54 \\
        GPT-4o-mini & 0.77 & 0.69 & - & - & - \\
        o1-mini & - & - & 0.27 & - & - \\
        TimeHC-RL (7B) & 0.80 & \textbf{0.80} & 0.35 & \textbf{0.60} & \textbf{0.70} \\
      \bottomrule
    \end{tabular}
  \end{adjustbox}
\end{table}



\end{document}